\documentclass[sigconf,nonacm]{acmart}
\AtBeginDocument{%
  }
\setcopyright{none}
\usepackage{booktabs}
\usepackage{multirow}
\usepackage{array}
\usepackage[table]{xcolor}      
\usepackage[normalem]{ulem}     
\usepackage{subcaption}
\usepackage{enumitem}

\usepackage{algorithm}
\usepackage{algpseudocode}
\usepackage{amsmath}

\usepackage[most]{tcolorbox}

\usepackage{graphicx}  

\begin{document}

\title{SAFE-G: Structure-aware Faithful Evidence-guided Generation for Knowledge-based Visual Question Answering}

\author{Long Shu}
\affiliation{%
  \institution{State Key Laboratory of Cognitive Intelligence, University of Science and Technology of China}
  \country{China}
}
\email{shulong@mail.ustc.edu.cn}

\author{Shuochen Liu}
\affiliation{%
  \institution{State Key Laboratory of Cognitive Intelligence, University of Science and Technology of China}
  \country{China}
}
\email{shuochenliu@mail.ustc.edu.cn}

\author{Wei Chen}
\affiliation{%
  \institution{State Key Laboratory of Cognitive Intelligence, University of Science and Technology of China}
  \country{China}
}
\email{chenweicw@mail.ustc.edu.cn}

\author{Junda Lin}
\affiliation{%
  \institution{State Key Laboratory of Cognitive Intelligence, University of Science and Technology of China}
  \country{China}
}
\email{linjunda@mail.ustc.edu.cn}

\author{Zhi Zheng}
\authornote{Corresponding authors}
\affiliation{%
  \institution{State Key Laboratory of Cognitive Intelligence, University of Science and Technology of China}
  \country{China}
}
\email{zhengzhi97@ustc.edu.cn}

\author{Huijun Hou}
\affiliation{%
  \institution{NIO}
  \country{China}
}
\email{houhj2009@163.com}

\author{Tong Xu}
\authornotemark[1]
\affiliation{%
  \institution{State Key Laboratory of Cognitive Intelligence, University of Science and Technology of China}
  \country{China}
}
\email{tongxu@ustc.edu.cn}

\renewcommand{\shortauthors}{Shu et al.}


\begin{abstract}
Knowledge-based Visual Question Answering (KB-VQA) aims to answer queries that necessitate reasoning over external knowledge sources beyond the visual content.
Typically, current methods fuse multimodal features to retrieve external information, subsequently leveraging Multimodal Large Language Models (MLLMs) to derive answers from the retrieved evidence.
However, these methods often struggle to capture structural associations within complex contexts to effectively filter noise. Furthermore, they frequently fail to ensure that the reasoning process remains strictly faithful to the retrieved evidence.
To address these challenges, we propose \textbf{\mbox{SAFE-G}}, a \textbf{S}tructure-\textbf{A}ware \textbf{F}aithful \textbf{E}vidence-guided \textbf{G}eneration framework, which enables precise evidence localization and trustworthy reasoning.
Specifically, we first employ a coarse-grained hybrid search fusing visual and textual modalities to recall candidate documents, and subsequently implement a structure-aware fine-grained graph retrieval that captures structural dependencies to filter noise and pinpoint precise evidence.
Moreover, we introduce a reinforcement learning (RL) strategy with an evidence-grounded reward that assigns credit to correct answers only when the selected evidence is correct. This strict alignment constraint compels the model to anchor its response in the retrieved context, effectively enhancing its capability to locate evidence via multimodal features and perform faithful reasoning.
Extensive experiments on the Encyclopedic-VQA and InfoSeek benchmarks demonstrate that SAFE-G outperforms prior methods by a margin of 8.9\% and 3.5\%, substantially enhancing the overall reasoning accuracy.
Our source code is publicly available at: \url{https://github.com/MINE-USTC/SAFE-G}.
\end{abstract}




\maketitle

\section{Introduction}

Visual Question Answering (VQA)~\cite{antol2015vqa, wu2017visual} aims to answer natural-language questions about images by reasoning over visual content. 
However, many real-world questions require knowledge that cannot be observed from pixels alone, such as domain expertise or contextual background information~\cite{deng2025comprehensive}. 
This need gives rise to Knowledge-based Visual Question Answering (KB-VQA)~\cite{marino2019ok, xenos2023simple}, where models must integrate visual understanding with external knowledge to produce accurate and well-grounded answers.

\begin{figure}[t] 
\centerline{
\includegraphics[width=1\linewidth]{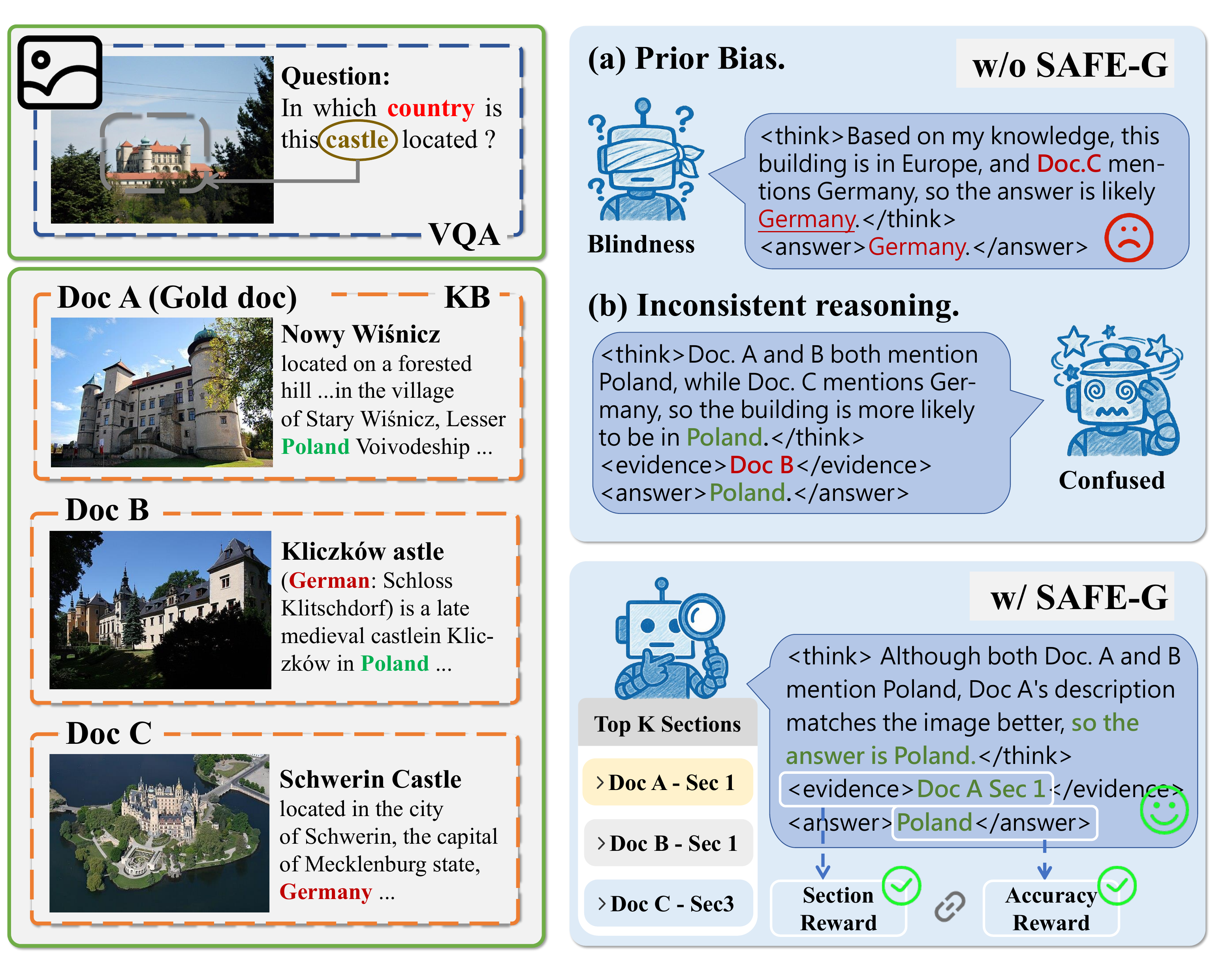} } 
\caption{ 
Comparison of reasoning behaviors under different reward schemes: (a) Optimizing solely for the final answer causes the model to overlook external documents. (b) Using separate rewards for retrieval and answering results in disconnected and inconsistent reasoning chains. (c) SAFE-G employs a holistic reward design that guides the model to learn a correct and coherent reasoning trajectory, ensuring the answer is faithfully anchored to the evidence.
} 
\label{fig:overview} 
\end{figure}

Early studies on KB-VQA typically combined dedicated retrievers with task-specific VQA architectures~\cite{lin2022retrieval, gao2022transform}, injecting external knowledge through handcrafted pipelines or modular reasoning components. 
With the advent of Multimodal Large Language Models (MLLMs), the dominant solution has shifted toward a Retrieval-Augmented Generation (RAG) paradigm~\cite{caffagni2024wiki, yan2024echosight, yang2025omgm}: relevant evidence is first retrieved from external sources and then fed into an MLLM to produce the final answer. 
By grounding generation on retrieved, potentially domain-specific information, such RAG-based KB-VQA systems can expand the model’s effective knowledge beyond its parameters.


Despite this progress, existing retrieval-augmented frameworks still face two fundamental limitations. 
(1) \textbf{Reliance on Expensive Re-ranker Training}. Existing fine-grained retrieval methods either rely on similarity-based passage retrieval or task-specific re-rankers trained on curated datasets~\cite{yan2024echosight, cocchi2025augmenting}. While the former treats knowledge entries as isolated units and struggles to aggregate complementary evidence across passages, the latter requires substantial data curation and training, limiting generalizability across domains.
(2) \textbf{Generation Disconnect and Prior Bias}. As illustrated in Fig.~\ref{fig:overview}, models may not consistently be compelled to ground their reasoning in the retrieved context. Without explicit alignment mechanisms, strong pre-trained priors may override external evidence, leading to hallucinations. Furthermore, even when context is accessed, the reasoning process often suffers from a disconnect, where the generated answer is not derived from the cited evidence. This disconnect results in unfaithful reasoning trajectories where the evidence fails to support the final answer.

To address these challenges, we present \textbf{SAFE-G}, a \underline{S}tructure-\underline{A}ware \underline{F}aithful \underline{E}vidence-guided \underline{G}eneration framework for KB-VQA, which is conceptually organized into two coupled components: \emph{structure-aware multimodal retrieval} and \emph{evidence-grounded generation}.
By explicitly coupling evidence selection with answer reasoning, SAFE-G produces knowledge-grounded answers, alleviating both retrieval and generation failures in existing RAG-based systems.
Specifically, SAFE-G first adopts a coarse-to-fine retrieval pipeline to suppress noise and progressively localize relevant evidence. 
It begins with a coarse-grained hybrid search that exploits multimodal information in the knowledge base to retrieve a small set of candidate articles. 
Building upon these candidates, we introduce a structure-aware refinement mechanism designed to achieve fine-grained evidence localization under a training-free paradigm. To this end, we construct a schema-less heterogeneous knowledge graph to interconnect disjointed section-level contexts. Crucially, we engineer a multimodal-guided propagation strategy: by injecting visual-semantic relevance into the graph topology, we direct the graph traversal to prioritize evidence that is not only textually coherent but also visually aligned with the query. This enables SAFE-G to explicitly navigate latent dependencies and aggregate complementary evidence without the need for training an expensive re-ranker.
During generation, to enforce factual consistency and evidence faithfulness, we introduce a reinforcement learning (RL) strategy based on Group Relative Policy Optimization (GRPO)~\cite{shao2024deepseekmath} with an evidence-grounded reward. 
Unlike outcome-based RL paradigms that optimize solely for final answer accuracy, our reward assigns credit only when the model both identifies correct supporting evidence and produces an accurate response.
This strict coupling trains MLLMs to anchor their reasoning in the retrieved context, suppressing reliance on unsupported priors and thereby mitigating hallucinations.
Our contributions are summarized as follows:

\begin{itemize}[leftmargin=*, topsep=3pt, itemsep=3pt]
\item We propose SAFE-G, a framework that aggregates relevant evidence via a fine-grained graph retrieval. This mechanism localizes relevant information without requiring a trainable re-ranker.
\item To ensure faithful generation, we introduce an evidence-grounded reinforcement learning strategy. Unlike outcome-based supervision, our method enforces strict adherence to the retrieved context by coupling answer correctness with evidence usage.
\item Extensive experiments on the Encyclopedic-VQA and InfoSeek benchmarks demonstrate that SAFE-G achieves state-of-the-art performance, validating the effectiveness of our structure-aware retrieval and evidence-grounded generation. 
\end{itemize}
    

\section{Related Work}
\subsection{Knowledge-based VQA}

Early research in Visual Question Answering (VQA) primarily focused on deriving answers from visual content~\cite{antol2015vqa}. However, to address queries requiring information beyond the image, the field has expanded to the KB-VQA task~\cite{wang2017fvqa, marino2019ok, schwenk2022okvqa, qi2024rora}.
While MLLMs~\cite{liu2024improved, bai2025qwen2} excel in general visual understanding, they struggle with knowledge-intensive queries requiring precise information beyond their pre-training~\cite{mensink2023encyclopedic, chen2023can}. To address the knowledge gap, Retrieval-Augmented Generation (RAG) has witnessed increasing adoption across the field.

Visual-language retrieval typically necessitates a cross-modal retrieval mechanism to effectively process and align multimodal queries with heterogeneous document sources. Methods like PreFLMR~\cite{lin2024preflmr} and MuKA~\cite{deng2025muka} enhance this process by optimizing fine-grained cross-modal interactions or leveraging object-level visual features to query external knowledge bases. Following a hierarchical or re-ranking paradigm, approaches such as Wiki-LLaVA~\cite{caffagni2024wiki} and EchoSight~\cite{yan2024echosight} first retrieve candidate documents via global similarity and then refine them using learned scorers to select relevant context. Subsequently, MR2AG~\cite{zhang2024mr} and ReflectiVA~\cite{cocchi2025augmenting} incorporate reflective mechanisms, enabling the model to explicitly assess the relevance of retrieved knowledge or its own generation process to improve accuracy.

Diverging from paradigms that rely on heavy training or learned re-rankers, SAFE-G introduces a multimodal structure-aware graph mechanism. By injecting visual-semantic signals into the graph initialization, our approach leverages training-free Personalized PageRank (PPR)~\cite{brin1998anatomy} to aggregate evidence that is both structurally central and visually consistent. This design enables the model to bridge isolated evidence fragments and filter noise through multimodal structural consistency, bypassing the dependency on expensive parametric patterns.

\begin{figure*}[t]
  \centering
  \includegraphics[width=\textwidth]{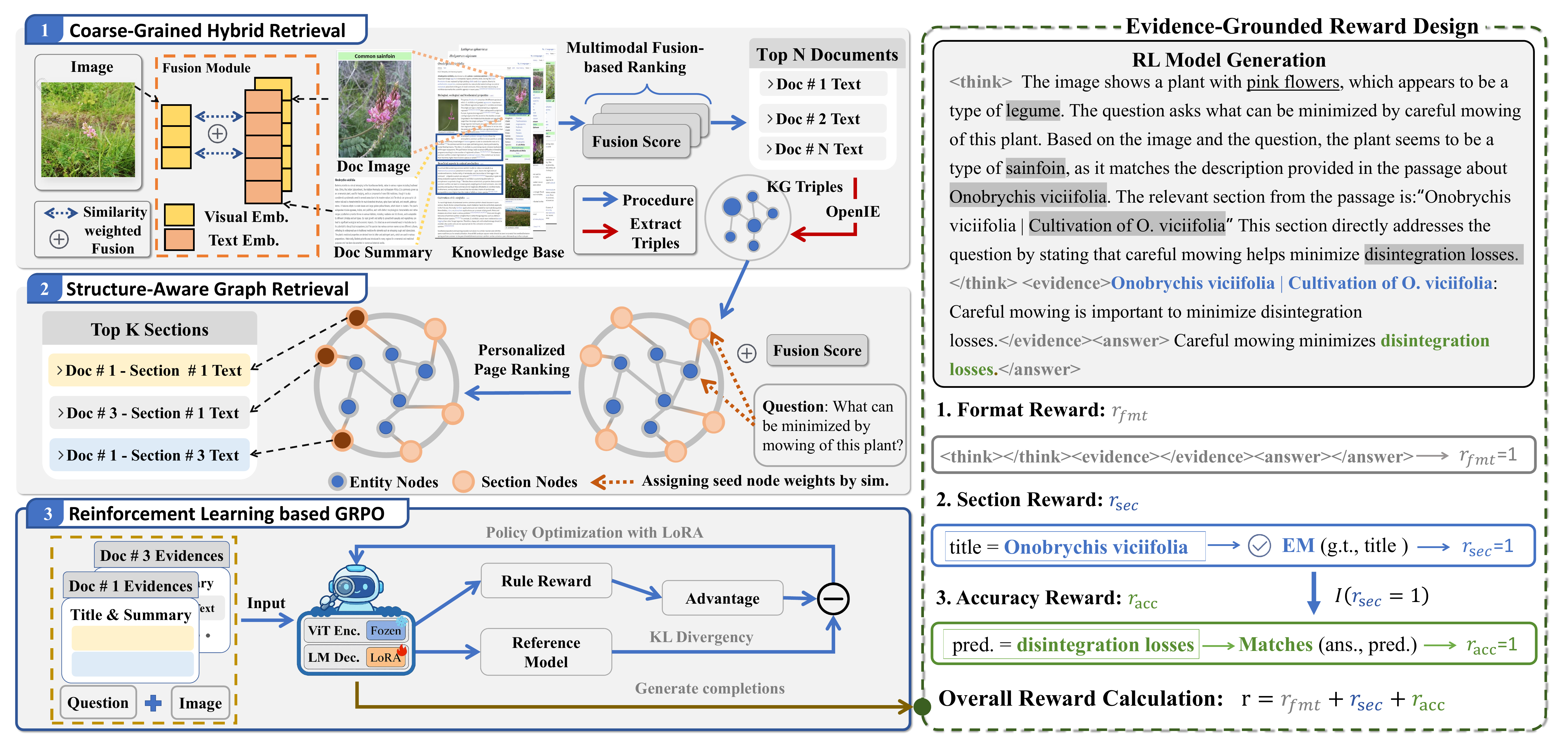}
  \caption{Overview of the SAFE-G framework. Left: A three-stage retrieval-augmented generation framework tailored for the KB-VQA task. Right: The Evidence-grounded Reward Design employed during GRPO training.}
  \label{fig:method}
\end{figure*}

\subsection{RL for MLLM Reasoning}
Recent research has shown that the alignment of Large Language Models (LLMs) has increasingly shifted toward RL-based policy optimization~\cite{zhang2024item}, which has proven effective in augmenting reasoning capabilities for complex problem-solving~\cite{jin2025search, song2025r1}. DeepSeek-R1~\cite{guo2025deepseek} marked a milestone, demonstrating that RL-driven post-training can elicit Chain-of-Thought (CoT) reasoning and trigger emergent ``aha moments''. 
Drawing from these textual advancements, recent initiatives have extended R1-style methodologies to Multimodal Large Language Models (MLLMs), employing rule-based incentives to enhance mathematical and perceptual reasoning~\cite{shen2025vlm, liu2025look}. Surpassing the limitations of conventional Supervised Fine-Tuning (SFT), RL paradigms enable more profound reasoning depth and generalization without the bottleneck of massive human labeling~\cite{chu2025sft}.

While alleviating the reliance on annotated data, these RL methods often shift the bottleneck to computational resources. To mitigate the computational burden, Group Relative Policy Optimization (GRPO)~\cite{shao2024deepseekmath} has emerged as a preferred, efficient alternative, distinguishing itself by eliminating the need for a concurrent value function critic. Within the specific landscape of KB-VQA, pioneering efforts such as VLM-PRF~\cite{hong2025knowledge} have successfully leveraged RL to optimize the efficiency of feedback integration, primarily utilizing outcome-level supervision to align model outputs with target answers. However, a critical challenge remains in ensuring that the generated answers are faithfully grounded in the retrieved context. To address this, our proposed SAFE-G introduces an evidence-grounded reward mechanism. By coupling the reward signal with the verification of supporting documents, SAFE-G compels the model to anchor its reasoning in retrieved evidence, thereby ensuring that reliability stems from verified context.

\section{Preliminaries}
Knowledge-based Visual Question Answering (KB-VQA) extends traditional visual reasoning by explicitly introducing an external knowledge retrieval constraint. Unlike standard VQA, where answers are derived directly from visual content, KB-VQA requires the model to answer a multimodal query by retrieving and reasoning over the relevant evidence from an explicit external knowledge base (e.g., Wikipedia).

We denote the external document collection as $\mathcal{D}=\{d_i\}_{i=1}^{N}$, where each entry $d_i$ corresponds to a Wikipedia page. 
Each document $d_i$ is associated with (i) metadata $M_i$ (e.g., document title and section title), (ii) the associated document image $I_i$, and (iii) a set of section-level textual units $\{s_{i,j}\}_{j=1}^{n_i}$. 
Here, $s_{i,j}$ denotes the $j$-th section of document $d_i$, where the section title and its content are treated as a single textual unit.

Given a query $q=(I_q, Q)$ consisting of an image $I_q$ and a question $Q$, KB-VQA can be formulated as retrieval-conditioned generation.
A retriever $\mathcal{R}$ first selects a query-specific subset of the document collection. Formally,
\begin{equation}
\mathcal{D}_q = \mathcal{R}(I_q, Q, \mathcal{D}), \quad \mathcal{D}_q \subseteq \mathcal{D},
\label{pre:doc}
\end{equation}
and MLLM parameterized by $\theta$ then predicts the final answer $y$ conditioned on the query and retrieved knowledge:
\begin{equation}
y = \mathcal{M}_{\theta}(I_q, Q, \mathcal{D}_q).
\end{equation}

We assess answer quality using BEM score~\cite{zhang2019bertscore} for E-VQA and VQA Accuracy~\cite{goyal2017making} for InfoSeek. Retrieval quality is measured by Recall@$K$ (R@$K$) at both document and section level.



\section{Method}

In this section, we present \textbf{SAFE-G}, a structure-aware and evidence-guided framework that jointly improves evidence retrieval and faithful generation.
As shown in Fig.~\ref{fig:method}, SAFE-G consists of three stages. 
In stage~1, we perform coarse-grained multimodal retrieval (\S\ref{sec:coarse}) to identify a set of candidate documents relevant to the query image. 
Then, we conduct fine-grained structure-aware graph retrieval (\S\ref{sec:fine}) over sections by constructing a query-specific graph and propagating relevance to aggregate relevant evidence across related sections as the second stage. 
In stage~3, we apply RL with an evidence-grounded objective to encourage the model to select evidence and generate answers that remain faithful to the selected context (\S\ref{sec:rl}). 
These components reduce retrieval noise, aggregate cross-passage evidence without re-ranker training, and improve the faithfulness of knowledge-grounded answering.

\subsection{Coarse-grained Multimodal Hybrid Retrieval}
\label{sec:coarse}
Directly performing fine-grained retrieval over the massive corpus $\mathcal{D}$ is inefficient, yet prior coarse-grained methods often suffer from single-modality limitations like visual ambiguity. To address this, we introduce a Coarse-grained Multimodal Hybrid Retrieval approach. By matching the query image against both document summaries and images, we leverage complementary multimodal signals to narrow the search space, ensuring high-quality candidates for the subsequent refinement.


\subsubsection{Summary Generation.}
A full Wikipedia document typically contains extensive details, which can introduce redundancy and noise when used directly as an indexing unit. Therefore, constructing a compact representation that retains only the salient information is crucial for robust retrieval. Following a summary-based indexing strategy~\cite{yang2025omgm}, we first generate a concise yet informative summary for each document to distill its core semantics, thereby aligning the textual evidence more closely with the query image.
Concretely, given the section-level texts $\{s_{i,j}\}_{j=1}^{n_i}$ of a document $d_i$, we feed the concatenated texts $a_i$ into a pre-trained summarization model $\mathcal{M}_s$ and apply an instruction prompt $P$ to produce the summary $m_i$, as follows:
\begin{align}
a_i &= \mathrm{Concat}\big(\{s_{i,j}\}_{j=1}^{n_i}\big), \label{eq:concat} \\
m_i &= \mathcal{M}_s(P, a_i). \label{eq:metadata}
\end{align}

\subsubsection{Hybrid Retrieval.}
After obtaining the generated document summaries, we perform hybrid retrieval conditioned on the query image $I_q$. Specifically, we project the query image, candidate document summaries, and associated document images into a shared feature space using CLIP encoders~\cite{radford2021learning}.
To capture both visual and semantic relevance, we compute similarity-based matching between the query image and (i) the candidate summary, (ii) the candidate image, and then combine these two signals via a weighted fusion to obtain the final relevance score for each document.
\begin{align}
Score_{d_i} &= \alpha \cdot \mathrm{sim}\big(E_v(I_q), E_t(m_i)\big) \nonumber \\
&\quad + (1-\alpha) \cdot \mathrm{sim}\big(E_v(I_q), E_v(I_i)\big), \label{eq:score} \\
\mathcal{C}_q &= \mathrm{Top}\text{-}K_d\Big(\{Score_{d_i}\}_{d_i \in \mathcal{D}}\Big), \label{eq:topk}
\end{align}
where $I_i$ is the image associated with document $d_i$, and $m_i$ denotes the document summary. $E_v(\cdot)$ and $E_t(\cdot)$ represent the visual and textual encoders, respectively. The hyperparameter $\alpha \in [0,1]$ balances the contribution of visual alignment versus semantic consistency.
For computational efficiency, we implement the retrieval using FAISS~\cite{johnson2019billion} to perform Maximum Inner-Product Search (MIPS) over the pre-indexed embeddings. Finally, we retrieve the top-$K_d$ documents according to $Score_{d_i}$ and denote the resulting candidate document set as $\mathcal{C}_q$.

\subsection{Structure-aware Graph Retrieval}
\label{sec:fine}

While the coarse-grained stage efficiently filters the corpus, answering complex visual queries requires identifying specific evidence segments hidden within the candidate documents. 
To achieve this, we perform Structure-aware Graph Retrieval to capture fine-grained semantic details within the candidate set. 
Inspired by HippoRAG~\cite{gutierrez2025rag}, we use  Open Information Extraction (OpenIE)\footnote{https://stanfordnlp.github.io/CoreNLP/openie.html} to construct graphs for the candidates $\mathcal{C}_q$. 
Crucially, we enhance this structure by injecting the multimodal relevance scores derived in the previous stage into the section nodes as a retrieval prior. By applying structure-aware propagation, we globally rank the sections and output the top-$K_s$ evidence units, which serve as the precise context for the subsequent generation stage.

\subsubsection{Graph Construction via OpenIE} 

Building on the candidate set $\mathcal{C}_q$, we construct a schema-less heterogeneous graph that encodes the structural dependencies of knowledge units.
For every section $s_{i,j}$, we extract relational triples $(h, r, t)$ via OpenIE,  where $h$ and $t$ denote the head and tail entities and $r$ denotes the relation between them. Then we define the graph $\mathcal{G}_q=(\mathcal{V}_q, \mathcal{E}_q)$ with section nodes and entity nodes as the vertex set.
The edge set $\mathcal{E}_q$ incorporates textual grounding by connecting section nodes to their mentioned entities (via \texttt{contain} edges), and relational structure by linking entities according to the extracted triples.
This topology links disjointed sections through shared entities, allowing subsequent propagation to aggregate evidence across documents.

\subsubsection{Multimodal-aware Section Prior}

While the graph $\mathcal{G}_q$ establishes structural pathways between sections, effectively traversing this structure requires a query-dependent initialization to guide the search. To this end, we construct a multimodal-aware prior by synergizing dense text matching with the multimodal relevance signals derived from Stage~1.

\paragraph{Section-node prior.}
To capture local textual relevance, we first compute a dense retrieval score $\mathrm{DPR}(s_{i,j})$ measuring the semantic correspondence between the question $Q$ and the section text.
However, looking at sections in isolation risks missing the broader visual context. To bridge this, we incorporate the global multimodal confidence from Stage~1.
Specifically, we re-calibrate the document-level score $Score_{d_i}$ via min-max normalization over the candidate set $\mathcal{C}_q$ to sharpen its discriminability:
\begin{equation}
\tilde{\rho}_i \;=\; 
\frac{Score_{d_i}-\min_{d_k \in \mathcal{C}_q}Score_{d_k}}
{\max_{d_k \in \mathcal{C}_q}Score_{d_k}-\min_{d_k \in \mathcal{C}_q}Score_{d_k}}.
\label{eq:stage2_weight}
\end{equation}
We then modulate the textual relevance with this multimodal prior to establish the final initial weight for each section node:
\begin{equation}
w_{i,j} \;=\; \mathrm{DPR}(s_{i,j}) \cdot \Big(1 + \beta \cdot \tilde{\rho}_i\Big),
\end{equation}
where $\beta \in [0,1]$ is a hyperparameter governing the influence of the global multimodal signal.



\paragraph{Entity-node initialization.}
To enable query-aware propagation over relational structures, we further activate relevant entity nodes in $\mathcal{G}_q$. 
Specifically, we compute semantic similarity scores between the question and the extracted OpenIE triples using the dense encoder.
Entities involved in these top-ranked triples accumulate the relevance scores, thereby being assigned higher initial weights and serving as auxiliary semantic seeds. 
Finally, we normalize the initial weights across all section and entity nodes in $\mathcal{G}_q$ to formulate the final personalization prior for graph propagation.


\subsubsection{Query-personalized Propagation and Evidence Selection}
With the constructed graph $\mathcal{G}_q$ and the multimodal-aware node prior, we perform query-personalized propagation to establish a global ranking over section nodes.
To implement this, we employ the Personalized PageRank (PPR) algorithm~\cite{brin1998anatomy}. Crucially, we directly utilize the multimodal prior derived in \S~4.2.2 as the personalization vector to initialize the random walk.
This setup directs the probability mass to diffuse from semantically and visually relevant anchors, propagating relevance through entity--relation connections to aggregate complementary cues across disjoint sections.
Upon convergence, the stationary distribution serves as the final relevance score. We then retain the top-$K_s$ ranked sections as the evidence set $\mathcal{E}_q$, passing them to the generation stage.


\subsection{Evidence-guided Generation via RL}
\label{sec:rl}
While the retrieval stage provides high-quality evidence $\mathcal{E}_q$, standard generation models may still hallucinate or ignore the provided context. To enforce strict faithfulness, we introduce a Reinforcement Learning (RL) method that couples evidence usage with answer correctness. This mechanism guides the model to anchor its reasoning process in the retrieved evidence, ensuring that the generated answer is  correct and grounded.

\subsubsection{Evidence-grounded     Reward Design}

To promote both answer accuracy and faithful evidence usage, we design a composite reward that evaluates the output based on three criteria: answer correctness, evidence selection, and evidence-grounded consistency.
We require the model to generate structured outputs containing its reasoning process in \texttt{<think>}, citations in \texttt{<evidence>}, and the final answer in \texttt{<answer>}.
This format allows us to parse the sampled output $o_i$ and extract both the predicted answer $\hat{y}_i$ and the cited content.
Since the evidence strings follow the specific format \texttt{Document Title | Section Title: snippet}, we can directly identify the predicted document title $\hat{T}_i$ and verify it against the ground-truth title $T^\ast$.

\paragraph{Section reward}
To guide the model to select knowledge in a targeted manner during reasoning, we introduce a section reward that encourages the chosen evidence to originate from the correct source document. 
This provides a lightweight supervision signal that promotes grounding without requiring strict section-level annotation. 
Formally, we define:
\begin{equation}
r_{\text{sec}}(o_i)=
\begin{cases}
1, & \text{if } \hat{T}_i = T^\ast,\\
0, & \text{otherwise},
\end{cases}
\label{eq:sec_reward}
\end{equation}
where $\hat{T}_i$ is the document title extracted from \texttt{<evidence>} and $T^\ast$ is the ground-truth supporting document title.

\paragraph{Accuracy reward}

We formulate an evidence-grounded accuracy reward to enforce faithful reasoning by strictly coupling answer correctness with evidence validity. Specifically, the model receives credit for a correct prediction if and only if its cited evidence aligns with the ground-truth source, as defined by gating the accuracy score with the section reward:
\begin{equation}
r_{\text{acc}}(o_i)=
\begin{cases}
1, & \text{if } \hat{y}_i \text{ matches } y \ \text{and}\ r_{\text{sec}}(o_i)=1,\\
0, & \text{otherwise},
\end{cases}
\label{eq:acc_reward}
\end{equation}
where $y$ is the ground-truth answer and the matching follows the benchmark evaluation protocol.

\paragraph{Format Reward}
To enforce strict compliance with the required output structure, we define a format reward:
\begin{equation}
r_{\text{fmt}}(o_i)=
\begin{cases}
1,  & \text{if the format is correct},\\
-1, & \text{otherwise}.
\end{cases}
\end{equation}
Penalizing outputs that violate the required format strengthens structural compliance and speeds up alignment to the desired template, allowing subsequent optimization to focus more effectively on evidence selection and answer quality.

\paragraph{Overall Reward}


Finally, we aggregate the components into the overall reward $r(o_i)$. We adopt a cascaded evaluation strategy where content quality is assessed only when the structural requirements are met:
\begin{equation}
r(o_i)=
\begin{cases}
    -1 & \text{if $r_{\text{fmt}}(o_i)$ = -1,} \\
    r_{\text{fmt}}(o_i) + r_{\text{sec}}(o_i) + r_{\text{acc}}(o_i) & \text{if $r_{\text{fmt}}(o_i)$ = 1.}
\end{cases}
\end{equation}
This formulation jointly enforces format compliance, incentivizes precise evidence selection, and ensures faithful answer generation by gating accuracy on evidence validity. Consequently, the policy is steered toward outputs that are not only correct but also strictly grounded in the retrieved knowledge, effectively mitigating ungrounded hallucinations.

\subsubsection{Training Algorithm}

We leverage Group Relative Policy Optimization (GRPO)~\cite{shao2024deepseekmath} to optimize the policy $\pi_\theta$ under our fine-grained reward constraints.
GRPO is particularly well-suited for this setting as it eliminates the need for a value function critic, instead estimating the baseline from the group average of sampled outputs.
By sampling a group of candidate responses per query and computing their group-wise relative advantages, GRPO iteratively updates $\pi_\theta$ to reinforce generation trajectories that yield superior performance relative to the group baseline.

\section{Experiments}

\subsection{Datasets and Evaluation Metrics}
\begin{table*}[t]
  \centering
  \caption{VQA accuracy on E-VQA and InfoSeek. Results for our method are highlighted in light blue. $\ast$ marks entries that are incomparable due to differences in knowledge bases.}
  \label{tab:main_results}
  \setlength{\tabcolsep}{3pt}
  \renewcommand{\arraystretch}{1.15}
  \begin{tabular}{l l l c c c c c}
    \toprule
    \multirow{2}{*}{\textbf{Method}} & \multirow{2}{*}{\textbf{Model}} & \multirow{2}{*}{\textbf{Retriever}}
    & \multicolumn{2}{c}{\textbf{E-VQA}} & \multicolumn{3}{c}{\textbf{InfoSeek}} \\
    \cmidrule(lr){4-5}\cmidrule(lr){6-8}
    & & & Single-Hop & All & Unseen-Q & Unseen-E & All \\
    \midrule

    \multicolumn{8}{c}{\textit{Zero-shot MLLMs}} \\
    \midrule
    BLIP-2~\cite{li2023blip}        & Flan-T5XL   & -- & 12.6 & 12.4 & 12.7 & 12.3 & 12.5 \\
    InstructBLIP~\cite{dai2023instructblip}  & Flan-T5XL   & -- & 11.9 & 12.0 &  8.9 &  7.4 &  8.1 \\
    LLaVA-v1.5~\cite{liu2024improved}    & Vicuna-7B   & -- & 16.3 & 16.9 &  9.6 &  9.4 &  9.5 \\
    Qwen2.5-VL-3B~\cite{bai2025qwen2}  & --          & -- & 17.9 & 19.6 & 20.4 & 21.9 &  21.4 \\
    Qwen2.5-VL-7B~\cite{bai2025qwen2}  & --          & -- & 21.7 & 20.3 & 22.8 & 24.1 &  23.7 \\
    \midrule

    \multicolumn{8}{c}{\textit{Retrieval-Augmented Models}} \\
    \midrule
    DPR\_V+T$^{\ast}$~\cite{lerner2024cross}   & Multi-passage BERT & CLIP ViT-B/32 & 29.1 & --   & --   & --   & 12.4 \\
    RORA-VLM$^{\ast}$~\cite{qi2024rora}   & Vicuna-7B          & CLIP + Google Search & -- & 20.3 & 25.1 & 27.3 & -- \\
    EchoSight$^{\ast}$~\cite{yan2024echosight}  & Mistral-7B/LLaMA-3-8B & EVA-CLIP-8B & 19.4 & -- & -- & -- & 27.7 \\
    Wiki-LLaVA~\cite{caffagni2024wiki} & Vicuna-7B          & CLIP ViT-L/14 + Contriever & 17.7 & 20.3 & 30.1 & 27.8 & 28.9 \\
    ReflectiVA~\cite{cocchi2025augmenting} & LLaMA-3.1-8B       & EVA-CLIP-8B & 28.0 & 29.2 & 40.4 & 39.8 & 40.1 \\
    MMKB-RAG~\cite{ling2025mmkb}   & Qwen2-7B           & EVA-CLIP-8B & 39.7 & 35.9 & 36.4 & 36.3 & 36.4 \\
    mKG-RAG~\cite{yuan2025mkg}   & LLaMA-3.1-8B           & QM-Retriever & 38.4 & 36.3 & 41.4 & 39.6 & 40.5 \\ 
    CC-VQA~\cite{hong2026cc} & Qwen2.5-VL-7B  & EVA-CLIP-8B  & \uline{41.4} & \uline{36.1} & \uline{44.7} & \uline{46.1} & \uline{45.1} \\
    \midrule
    
    \multicolumn{8}{c}{\textit{Retrieval-Augmented Models with RL}} \\
    \midrule
    VLM-PRF~\cite{hong2025knowledge}    & Qwen2.5-VL-3B      & EVA-CLIP-8B & 31.1 & 32.4 & 39.7 & 38.8 & 39.0 \\
    
    \rowcolor{blue!5}
    \textbf{SAFE-G~(Ours)}       & Qwen2.5-VL-3B      & EVA-CLIP-8B
               & 39.2 & 37.9 & 40.9 & 40.8 & 40.9 \\

    VLM-PRF~\cite{hong2025knowledge}    & Qwen2.5-VL-7B      & EVA-CLIP-8B
               & 37.1 & 36.0 & 43.3 & 42.7 & 42.8 \\
    VLM-PRF~\cite{hong2025knowledge}    & InternVL3-8B       & EVA-CLIP-8B
               & 40.1 & 39.2 & 43.5 & 42.1 & 42.5 \\

    \rowcolor{blue!5}
    \textbf{SAFE-G~(Ours)}       & Qwen2.5-VL-7B      & EVA-CLIP-8B
               & \textbf{45.1} & \textbf{44.3} & \textbf{46.3} & \textbf{47.1} & \textbf{46.7} \\
    \bottomrule
  \end{tabular}
\end{table*}

\paragraph{Encyclopedic-VQA}
The Encyclopedic-VQA (E-VQA)~\cite{mensink2023encyclopedic} dataset is a KB-VQA benchmark that targets questions about fine-grained entities. It provides 221K distinct question–answer pairs, and each question is paired with up to five images (drawn from iNaturalist~\cite{van2021benchmarking} and Google Landmarks v2~\cite{weyand2020google}). The data are organized into train/validation/test splits with approximately 1M / 13.6K / 5.8K instances, and we follow prior work~\cite{yan2024echosight, hong2025knowledge} by reporting performance on the 5.8K test set. To support retrieval-based methods, E-VQA also releases a large Wikipedia-derived document collection (about 2M pages), where each page contains the document title, sectioned text, and associated images; we use this original collection in our experiments.
\paragraph{InfoSeek}
The InfoSeek~\cite{chen2023can} dataset is a large-scale benchmark for visual information-seeking questions. It contains about 1.3M image--question--answer triplets aligned with roughly 11K Wikipedia entities/pages. The dataset is split into train/validation/test sets of approximately 934K/73K/348K samples. InfoSeek also provides a Wikipedia-based knowledge base with 6M Wikipedia entries. Following common practice in prior work~\cite{yan2024echosight, hong2025knowledge}, we conduct retrieval over a 100K-page subset to balance coverage and efficiency.
\paragraph{Evaluation Metrics}
We adopt the official evaluation protocols released with each benchmark. 
For E-VQA, we use the BERT-based Matching (BEM) score~\cite{zhang2019bertscore} to evaluate answer quality by measuring the semantic similarity between the predicted and ground-truth answers. 
For InfoSeek, the evaluation follows its standard setting: we report VQA Accuracy~\cite{goyal2017making, marino2019ok}, which provides a more tolerant matching criterion for open-ended information-seeking answers.

\subsection{Implementation Details}
\paragraph{Retrieval Details}
For coarse-grained retrieval, to ensure consistency in the retrieval setting, we utilize the open-source document summaries provided by OMGM~\cite{yang2025omgm} and encode images via EVA-CLIP-8B~\cite{sun2024eva}. We perform FAISS-based retrieval in two steps: initially filtering the top-40 documents via summary matching, then re-ranking to select the top-20 candidates using the fused score (Eq.~\ref{eq:topk}). In the fine-grained stage, we employ LLaMa-3.3-70B-Instruct~\cite{grattafiori2024llama} for OpenIE extraction. Separately, we adopt \texttt{nvidia/NVEmbed-v2}~\cite{lee2024nv} as the unified encoder for all dense matching tasks, including question--section and query--triple relevance.

\paragraph{Generator Training Details}
We utilize Qwen2.5-VL-3B/7B~\cite{bai2025qwen2} as our backbone generators and employ LoRA~\cite{hu2022lora} ($r=64$, $\alpha=64$) for parameter-efficient fine-tuning, keeping the vision transformer (ViT) frozen. The GRPO optimization uses a learning rate of $1\times10^{-5}$ with cosine decay, a micro-batch size of 2 with 2 gradient accumulation steps, and $G=8$ rollouts per query at temperature 0.9. The maximum generation length is set to 600 tokens, with maximum prompt lengths of 16,384 and 32,768 tokens for E-VQA and InfoSeek, respectively. We sample 4K instances from the E-VQA training split and 4K instances from ReflectiVA~\cite{cocchi2025augmenting} for InfoSeek. All experiments are conducted on 8 NVIDIA RTX A6000 (48GB) GPUs, taking 24 hours per model, with DeepSpeed~\cite{rasley2020deepspeed} ZeRO-Offload~\cite{rajbhandari2021zero} and Flash-Attention~2~\cite{dao2023flashattention} for memory and throughput optimization.

\subsection{Main Results}

\subsubsection{VQA Results}
We conduct a comprehensive evaluation of SAFE-G on E-VQA and InfoSeek, comparing it against a broad set of baselines that cover zero-shot and retrieval-augmented baselines. For zero-shot performance, where MLLMs rely solely on the query inputs, we evaluated BLIP-2~\cite{li2023blip}, InstructBLIP~\cite{dai2023instructblip}, LLaVAv1.5~\cite{liu2024improved}, and Qwen2.5-VL~\cite{bai2025qwen2}. For retrieval-augmented methods, we compare frameworks including EchoSight~\cite{yan2024echosight}, WikiLLaVA~\cite{caffagni2024wiki}, MMKB-RAG~\cite{ling2025mmkb}, mKG-RAG ~\cite{yuan2025mkg}, ReflectiVA~\cite{cocchi2025augmenting}, and CC-VQA~\cite{hong2026cc}.
We include VLM-PRF~\cite{hong2025knowledge}, an approach that utilizes reinforcement learning for optimization.

As detailed in Tab.~\ref{tab:main_results}, zero-shot MLLMs struggle on both benchmarks, constrained by the static knowledge within their pre-trained parameters. This underscores the necessity of external retrieval for open-ended visual reasoning. While standard RAG baselines provide a performance lift, they often lack the mechanism to align generation strictly with the retrieved context. 
Recent RL-based approaches like VLM-PRF have attempted to bridge this gap. However, they primarily optimize for final answer correctness, often overlooking the intermediate reasoning quality. Consequently, models may still fail to filter irrelevant noise or hallucinate despite having correct evidence. SAFE-G addresses these limitations by enforcing structure-aware filtering and evidence-grounded faithfulness. By penalizing ungrounded answers through our composite reward, we compel the model to anchor its reasoning path in verified knowledge sections rather than treating retrieval as a black box.


In our experiments, SAFE-G achieves substantial gains across both benchmarks. On E-VQA, SAFE-G with Qwen2.5-VL-7B achieves 45.1\%, surpassing the prior SOTA CC-VQA by \textbf{3.7} points and VLM-PRF by \textbf{+8.0} points. Although CC-VQA also employs a strong Qwen2.5-VL-7B backbone, it lacks an explicit evidence-grounded generation mechanism, confirming that retrieval quality alone is insufficient without faithful reasoning. The improvement is even more pronounced on the smaller 3B model; notably, SAFE-G-3B (39.2\%) even surpasses VLM-PRF-7B (37.1\%), demonstrating that fine-grained evidence supervision effectively compensates for limited model capacity. On InfoSeek, SAFE-G yields consistent improvements over both CC-VQA (\textbf{+1.6} points) and VLM-PRF (\textbf{+1.9} and \textbf{+3.9} points), setting a new SOTA across both model scales.

\subsubsection{Results with Oracle Documents.}

\begin{table}[t]
  \centering
  \caption{VQA accuracy scores on E-VQA and InfoSeek with oracle Wikipedia pages.}
  \label{tab:oracle}
  \setlength{\tabcolsep}{4pt}
  \renewcommand{\arraystretch}{1.1}
  \resizebox{\linewidth}{!}{
  \begin{tabular}{l l c c c c}
    \toprule
    \multirow{2}{*}{\textbf{Method}} & \multirow{2}{*}{\textbf{Generator}} & \multicolumn{1}{c}{\textbf{E-VQA}} & \multicolumn{3}{c}{\textbf{Infoseek}} \\
    \cmidrule(lr){3-3} \cmidrule(lr){4-6}
     &  & Single-Hop & Un-Q & Un-E & All \\
    \midrule
    \rowcolor{blue!5}
    Base model & Qwen2.5-VL-7B & 35.7 & 39.1 & 38.0 & 35.2 \\

    \midrule
    ReflectiVA~\cite{cocchi2025augmenting} & Qwen2.5-VL-7B & 71.3 & 56.1 & 55.9 & 56.0 \\
    \rowcolor{blue!5}
    \textbf{SAFE-G~(Ours)}    & Qwen2.5-VL-3B & 63.1 & 56.7 & 55.5 & 56.1 \\

    \midrule
    Wiki-LLaVA~\cite{caffagni2024wiki} & LLaMA-3.1-8B & 46.8 & 51.2 & 50.6 & 50.9 \\
    ReflectiVA~\cite{cocchi2025augmenting} & LLaMA-3.1-8B & \uline{75.2} & \uline{57.8} & \uline{57.4} & \uline{57.6} \\
    ReflectiVA~\cite{cocchi2025augmenting} & Qwen2.5-VL-7B & 72.9 & 53.4 & 53.9 & 53.7 \\
    \rowcolor{blue!5}
    \textbf{SAFE-G~(Ours)}    & Qwen2.5-VL-7B & \textbf{82.8} & \textbf{60.0} & \textbf{58.8} & \textbf{59.4} \\
    \bottomrule
  \end{tabular}}
\end{table}

    
To estimate the upper bound of SAFE-G, we follow prior work~\cite{cocchi2025augmenting} and evaluate under an oracle setting, where the ground-truth Wikipedia page associated with each query is provided. We compared (i) a zero-shot MLLM baseline that directly consumes the full oracle document as input, and (ii) retrieval-augmented pipelines that further apply model-specific filtering over the oracle page (e.g., passage selection) before feeding the resulting context to the generator.
As shown in Tab.~\ref{tab:oracle}, simply providing the oracle document is insufficient: the Zero-shot baseline yields only 35.7\% on E-VQA, lagging behind SAFE-G by over \textbf{47} points. This massive gap indicates that standard MLLMs struggle to filter noise within long contexts. In contrast, SAFE-G benefits from its fine-grained graph refinement and RL-based optimization. 
Even in this high-information density setting, our method effectively localizes the precise evidence and enforces faithful reasoning, thereby substantially raising the achievable performance ceiling compared to prior best methods. Notably, under the same Qwen2.5-VL-7B generator, SAFE-G outperforms ReflectiVA by \textbf{9.9} points (82.8\% vs.\ 72.9\%), demonstrating that the performance gap stems not from model capacity but from superior evidence localization and faithful reasoning — capabilities that our graph refinement and evidence-gated RL are specifically designed to address.


\subsubsection{Coarse-grained Hybrid Retrieval Results}

    

As the entry point of our multi-stage pipeline, Stage~1 aims to narrow the million-scale corpus down to a manageable candidate set visually aligned with the query entity in $I_q$. 
To validate the impact of different retrieval signals, we compared several variants on E-VQA and InfoSeek (Tab.~\ref{tab:retrieval_two_datasets}).
Quantitatively, we find that relying on a single modality is suboptimal for coarse filtering. We observe that both \textit{Image$\rightarrow$Document} and \textit{Image$\rightarrow$Image} achieve limited recall, suggesting that neither page-level textual matching nor pure visual similarity alone can capture the full cues. While retrieving against high-information-density summaries proves more effective than using full documents, single-modality approaches still struggle to cover all critical features, leading to inferior alignment.
In contrast, our hybrid retriever consistently improves R@K on both datasets. By synergizing image--image and image--summary matching, our method exploits complementary multimodal signals, yielding a notably higher-quality candidate set $\mathcal{C}_q$ for the subsequent structure-aware refinement.


\begin{table}[t]
  \centering
  \caption{Comparison of coarse-grained retrieval performance (R@K) on E-VQA and Infoseek datasets.}
  \label{tab:retrieval_two_datasets}
  \setlength{\tabcolsep}{3pt} 
  \begin{tabular}{lcccccccc}
    \toprule
    \multirow{2}{*}{\textbf{Method}} & \multicolumn{4}{c}{\textbf{E-VQA}} & \multicolumn{4}{c}{\textbf{Infoseek}} \\
    \cmidrule(lr){2-5} \cmidrule(lr){6-9}
    & R@1 & R@5 & R@10 & R@20 & R@1 & R@5 & R@10 & R@20 \\
    \midrule
    I$\to$Doc. & 13.2 & 27.7 & 35.5 & 41.7 & 45.6 & 67.1 & 73.0 & 77.9 \\
    I$\to$Img. & 13.4 & 31.8 & 41.9 & 48.8 & 43.7 & 64.9 & 72.8 & 79.6 \\
    I$\to$Sum. & 19.1 & 41.2 & 49.8 & 58.7 & 52.6 & 73.9 & 80.0 & 84.8 \\
    \rowcolor{blue!5} 
     \textbf{I$\to$Hyb.} & \textbf{25.6} & \textbf{44.6} & \textbf{53.7} & \textbf{59.7} & \textbf{56.7} & \textbf{76.0} & \textbf{81.7} & \textbf{85.6} \\
    \bottomrule
  \end{tabular}
\end{table}

\subsubsection{Fine-grained Graph Retrieval Results}
\begin{table}[t]
  \centering
  \caption{Fine-grained section retrieval performance on E-VQA. ``S.~R@k'' denotes the recall of the top-$k$ sections.}
  \label{tab:stage2_retrieval}
  \setlength{\tabcolsep}{3.5pt}
  \renewcommand{\arraystretch}{1.1}
  \resizebox{\linewidth}{!}{
  \begin{tabular}{l c c c c}
    \toprule
    \textbf{Method} & S.~R@1 &  S.~R@5 &  S.~R@10 &  S.~R@20 \\
    \midrule
    EchoSight \emph{w/o} train        & 0.0301 & 0.1036 & 0.1775 & 0.2901 \\
    EchoSight                 & \uline{0.2429} & \textbf{0.4034} & \uline{0.4217} & \textbf{0.4423} \\

    \rowcolor{blue!5}
    \textbf{SAFE-G~(Ours)}             & \textbf{0.2530} & \uline{0.4020} & \textbf{0.4230} & \uline{0.4407} \\
    \bottomrule
  \end{tabular}}
\end{table}

\begin{table*}[t]
  \centering
  \caption{Ablation study results on E-VQA and InfoSeek to validate the effectiveness of our model component. \checkmark indicates the inclusion of a specific module.}
  \label{tab:ablation_final}
  \setlength{\tabcolsep}{5pt}
  \renewcommand{\arraystretch}{1.15}
  \begin{tabular}{l c c c c c c c c c}
    \toprule
    \multirow{2}{*}{\textbf{Model}} & \multicolumn{2}{c}{\textbf{Graph Retrieval}} & \multicolumn{3}{c}{\textbf{Reinforcement Learning}} & \textbf{E-VQA} & \multicolumn{3}{c}{\textbf{InfoSeek}} \\
    \cmidrule(lr){2-3} \cmidrule(lr){4-6} \cmidrule(lr){7-7} \cmidrule(lr){8-10}
    & PPR & w/~image & w/~evidence & w/~gating & w/~summary & Single-hop & Unseen-Q & Unseen-E & All \\
    \midrule

    \multirow{4}{*}{Qwen2.5-VL-7B} & \checkmark & & \checkmark & \checkmark & \checkmark & 42.7 & 43.2 & 44.6 & 43.5 \\
    
    & \checkmark & \checkmark & & & & 34.8 & 33.8 & 35.4 & 34.9 \\
    
    & \checkmark & \checkmark & \checkmark & & & 38.3 & 41.5 & 41.9 & 41.7 \\
    
    & \checkmark & \checkmark & \checkmark & \checkmark & & 43.2 & 44.7 & 45.3 & 45.1 \\
    
    \midrule
    \rowcolor{blue!5}
    \textbf{SAFE-G(Ours)} 
    & \checkmark & \checkmark & \checkmark & \checkmark & \checkmark & \textbf{45.1} & \textbf{46.3} & \textbf{47.1} & \textbf{46.7} \\
    
    \bottomrule
  \end{tabular}
\end{table*}

To compare SAFE-G's structure-aware graph retrieval with the learning-based reranking approach EchoSight~\cite{yan2024echosight}, which trains a task-specific Q-Former reranker on curated data, we evaluated section-level recall on the single-hop split of E-VQA under the same Stage~1 candidate pool (document Recall@20 = 44.5\%).
As shown in Tab.~\ref{tab:stage2_retrieval}, applying EchoSight's Q-Former directly without task-specific fine-tuning (\emph{w/o train}) yields drastically lower recall, confirming that learning-based re-rankers rely heavily on in-domain supervision to generalize. In contrast, SAFE-G achieves comparable section retrieval performance to the fully trained EchoSight model solely through PPR-based propagation, demonstrating that competitive fine-grained evidence localization can be achieved without expensive data construction or model training.


\subsection{Ablation Study}

We conduct a comprehensive ablation study on both E-VQA and InfoSeek to validate the contribution of each component in SAFE-G. As presented in Tab.~\ref{tab:ablation_final}, we progressively remove individual modules from the full model, and the results are analyzed as follows.

\textbf{(1) Multimodal Prior Drives Evidence Localization.} Removing the image-guided node initialization from graph propagation (row~1) reduces VQA accuracy to 42.7\% on E-VQA, a drop of 2.4 points from the full model. This confirms that injecting visual-semantic relevance into the graph initialization is essential: without this multimodal prior, the propagation relies solely on text-based structural connectivity, which is insufficient to align evidence localization with the visual query context in KB-VQA.

\textbf{(2) Evidence-grounded Reward Enforces Faithful Reasoning.} When the base model directly consumes graph-retrieved evidence without any RL training (row~2), performance stands at 34.8\%, confirming that high-quality retrieval alone is insufficient without faithful generation. Adding the evidence-grounded reward (row~3) lifts performance to 38.3\%, showing that explicitly coupling evidence selection with answer correctness teaches the model to ground its reasoning in the retrieved context rather than relying on parametric priors.

\textbf{(3) Evidence Gating Prevents Ungrounded Reward.} Adding the evidence gating (row~4) further improves performance from 38.3\% to 43.2\%. Without this mechanism, the accuracy reward is granted regardless of evidence correctness, allowing the model to receive positive feedback from parametrically correct answers without grounding in the retrieved context. By conditioning $r_{\text{acc}}$ on $r_{\text{sec}}=1$, this gating penalizes ungrounded answers and compels the model to cite correct evidence before receiving accuracy credit.

\textbf{(4) Summaries Align Retrieval with Visual Queries.} The full SAFE-G model achieves 45.1\%, outperforming the w/o summary variant (43.2\%) by 1.9 points. Document summaries provide compact representations that align the coarse-grained multimodal retrieval signal more closely with the visual query, yielding consistent gains across all evaluation splits on both benchmarks.

\subsection{Further Analysis}
\subsubsection{Effect of Fusion Weight $\alpha$ in Coarse-grained Retrieval}

\begin{figure}[t]
  \centering
  \includegraphics[width=\linewidth]{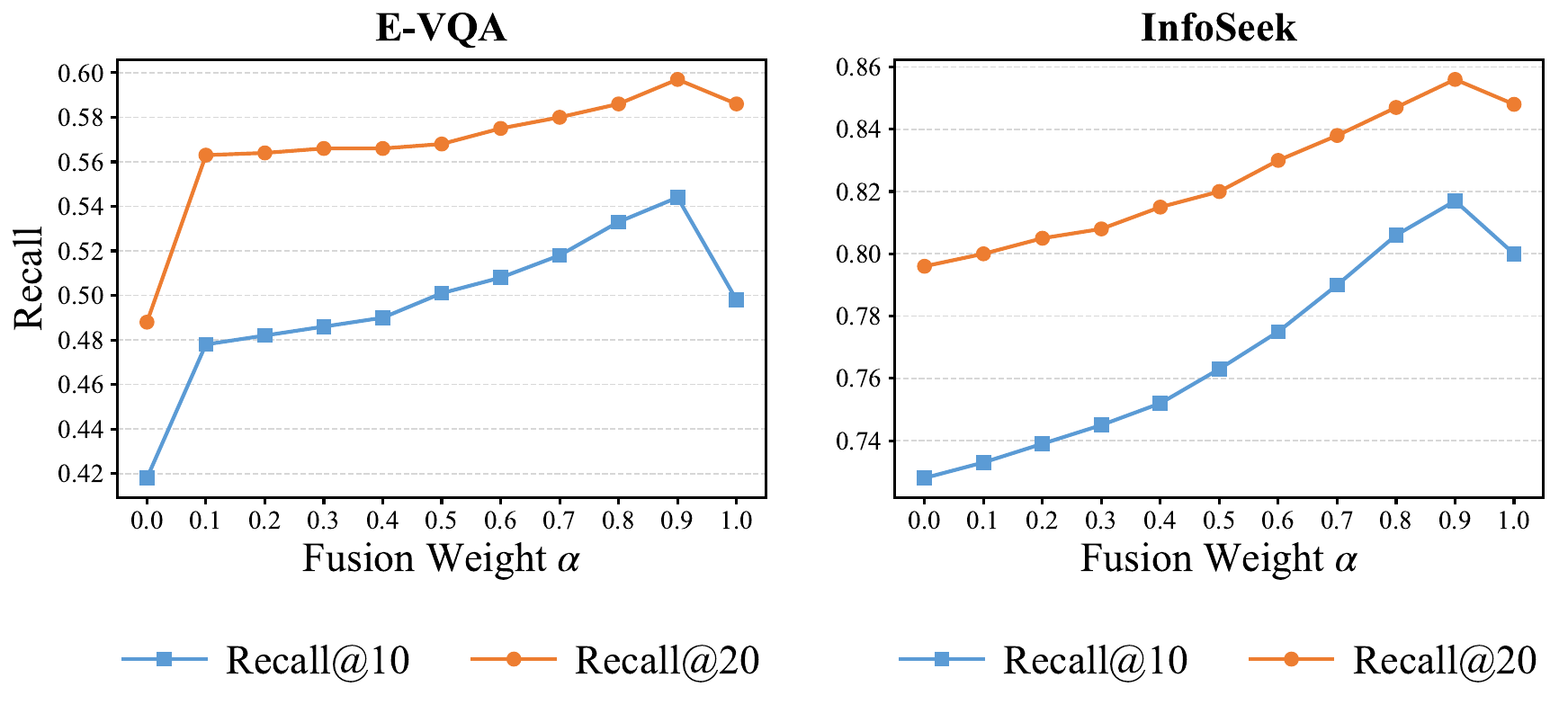}
  \caption{Retrieval performance (Recall@10/20) under different $\alpha$ settings on E-VQA and Infoseek datasets.}
  \label{fig:alpha}
\end{figure}

\begin{figure*}[ht]
  \centering
  \includegraphics[width=\textwidth]{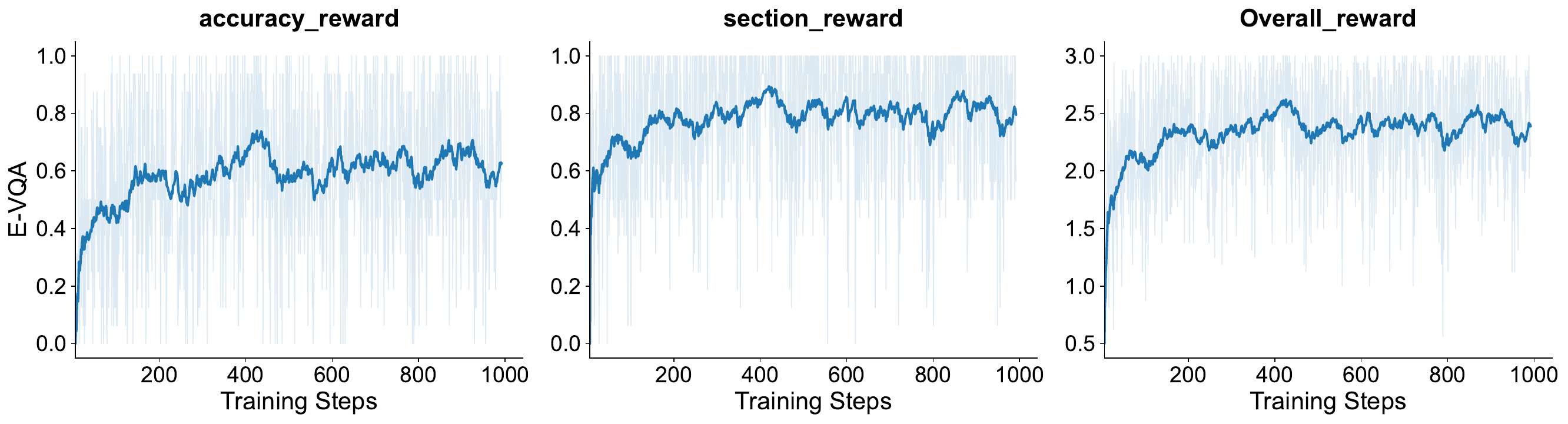}
  \caption{Training curves of SAFE-G-7B on E-VQA, showing the trajectories of $r_{\text{acc}}$, $r_{\text{sec}}$, and overall reward $r$. All metrics increase synchronously and stabilize upon convergence, validating the effectiveness of our evidence-gated reward design.}
  \label{fig:app_reward}
\end{figure*}

In this section, we explored the impact of the fusion weight $\alpha$ in Eq.~(\ref{eq:topk}) on the performance of coarse-grained retrieval. As shown in Fig.~\ref{fig:alpha}, relying solely on one modality proves suboptimal in large-scale search tasks: pure image retrieval ($\alpha=0$) is often distracted by visual ambiguities (e.g., entities with similar appearances or backgrounds), while pure summary-based retrieval ($\alpha=1$) may struggle with semantic confusion, especially when multiple candidates share overlapping textual descriptions.

When $\alpha$ is small, even a modest amount of summary matching added to image-only retrieval provides a noticeable performance boost, demonstrating that textual cues help disambiguate visually similar entities. As $\alpha$ increases, the performance generally continues to improve, indicating that summary-level semantics provide a strong signal for identifying the correct entity, while the image modality remains essential in aligning entities. Notably, the best performance is observed at $\alpha=0.9$, with a slight decline when moving to the text-only setting ($\alpha=1$). This suggests that while summary matching is the dominant factor, retaining a small contribution from image similarity is beneficial for robust retrieval. Therefore, we set $\alpha=0.9$ in all experiments unless otherwise stated.

\subsubsection{Results on more benchmarks}
\label{okvqa}

\begin{table}[t]
  \centering
  \caption{VQA accuracy on the OK-VQA benchmark. SAFE-G is evaluated at both 3B and 7B model scales and compared against zero-shot, RAG, and RL-based baselines.}
  \label{tab:okvqa}
  \setlength{\tabcolsep}{4pt}
  \renewcommand{\arraystretch}{1.1}
  \begin{tabular}{l l c c}
    \toprule
    \textbf{Method} & \textbf{Retriever} & \textbf{Model} & \textbf{OK-VQA} \\
    \midrule
    
    \rowcolor{blue!5}
    Base model & -- & Qwen2.5-VL-7B & 64.7 \\
    \midrule
    
    KU-RAG~\cite{zhang2025fine} & -- & LLaVA-Next-7B & 73.1 \\
    MMKB-RAG~\cite{ling2025mmkb} & PreFLMR & LLaMA-3.1-8B & 65.4 \\
    \midrule
    
    VLM-PRF~\cite{hong2025knowledge} & EVA-CLIP-8B & Qwen2.5-VL-3B & 68.6 \\
    \rowcolor{blue!5}
    \textbf{SAFE-G~(Ours)} & EVA-CLIP-8B & Qwen2.5-VL-3B & 70.2 \\
    \midrule
    
    VLM-PRF~\cite{hong2025knowledge} & EVA-CLIP-8B & Qwen2.5-VL-7B & \uline{77.8} \\
    \rowcolor{blue!5}
    \textbf{SAFE-G~(Ours)} & EVA-CLIP-8B & Qwen2.5-VL-7B & \textbf{79.8} \\
    
    \bottomrule
  \end{tabular}
\end{table}

To further validate the generalizability of SAFE-G beyond the E-VQA and InfoSeek benchmarks, we evaluate on OK-VQA~\cite{marino2019ok}, a widely adopted benchmark requiring external world knowledge. As shown in Tab.~\ref{tab:okvqa}, SAFE-G consistently outperforms all baselines at both model scales. With Qwen2.5-VL-7B, SAFE-G achieves 79.8\%, surpassing the strongest baseline VLM-PRF by \textbf{2.0} points and outperforming the zero-shot base model by a large margin of \textbf{15.1} points. With the smaller 3B backbone, SAFE-G still achieves 70.2\%, exceeding VLM-PRF by \textbf{1.6} points. These results demonstrate that SAFE-G's structure-aware retrieval and evidence-grounded generation strategy transfer effectively to a different knowledge-intensive benchmark, confirming the broad applicability of the proposed framework.

\subsubsection{Efficiency Analysis}

SAFE-G achieves competitive inference efficiency while requiring no model training. At query time, the online pipeline consists of coarse retrieval (3.67s) and PPR-based graph propagation (1.03s), totaling 4.70s per query — approximately 34\% higher than a representative trainable RAG baseline. However, this modest overhead is justified by the elimination of training costs: methods such as EchoSight~\cite{yan2024echosight} require constructing curated contrastive training data and approximately 40 GPU-hours of fine-tuning before deployment. The offline preprocessing costs of SAFE-G (OpenIE extraction and graph indexing) are amortized over the corpus before serving and do not affect query-time latency. All components are built upon open-source models, making the full pipeline locally deployable without proprietary API dependencies or external service cost.

\subsubsection{Effectiveness of Reinforcement Learning.}

\begin{figure*}[t]
  \centering
  \begin{subfigure}{0.515\textwidth}
    \includegraphics[width=\linewidth]{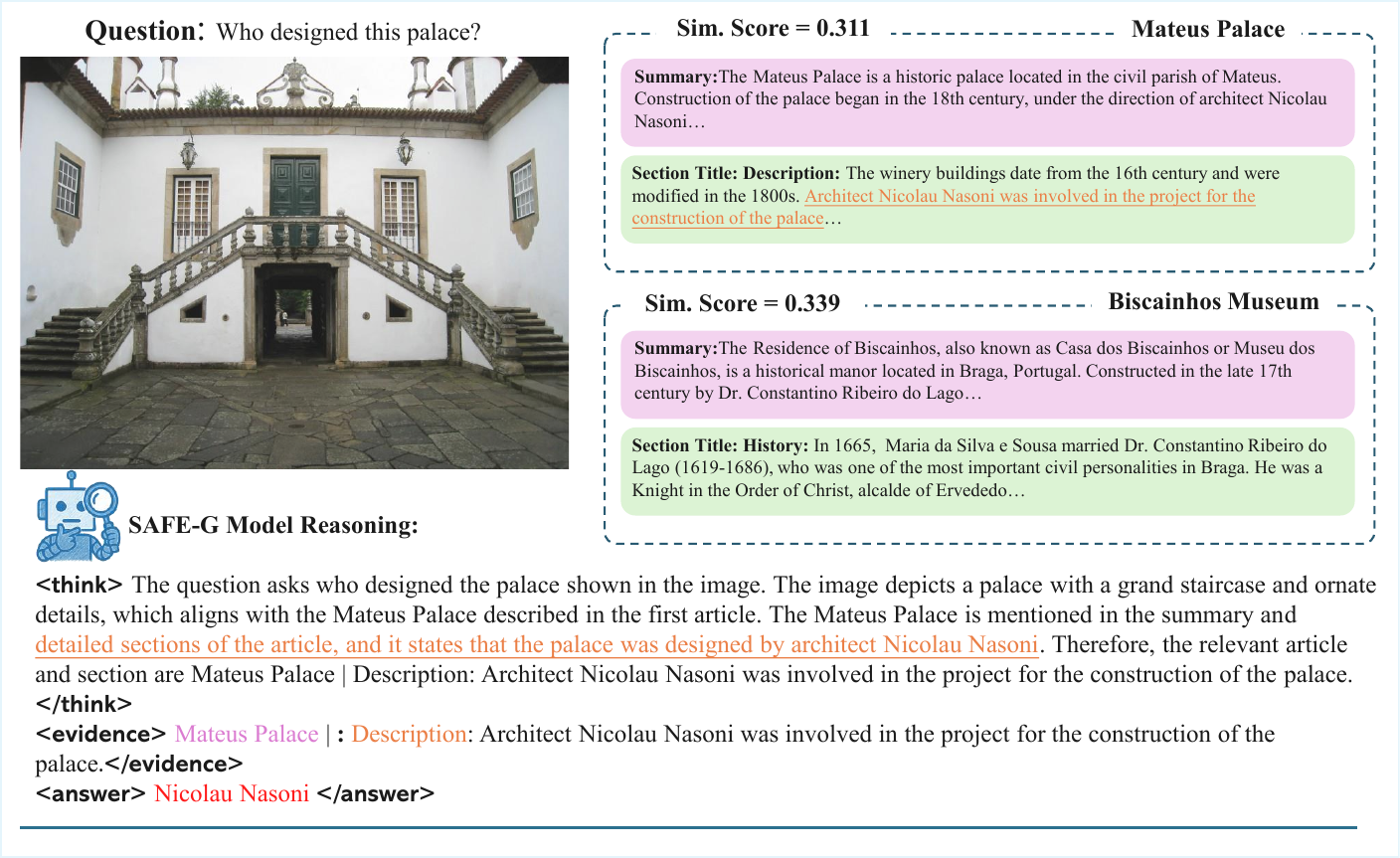}
  \end{subfigure}
  \hfill
  \begin{subfigure}{0.48\textwidth}
    \includegraphics[width=\linewidth]{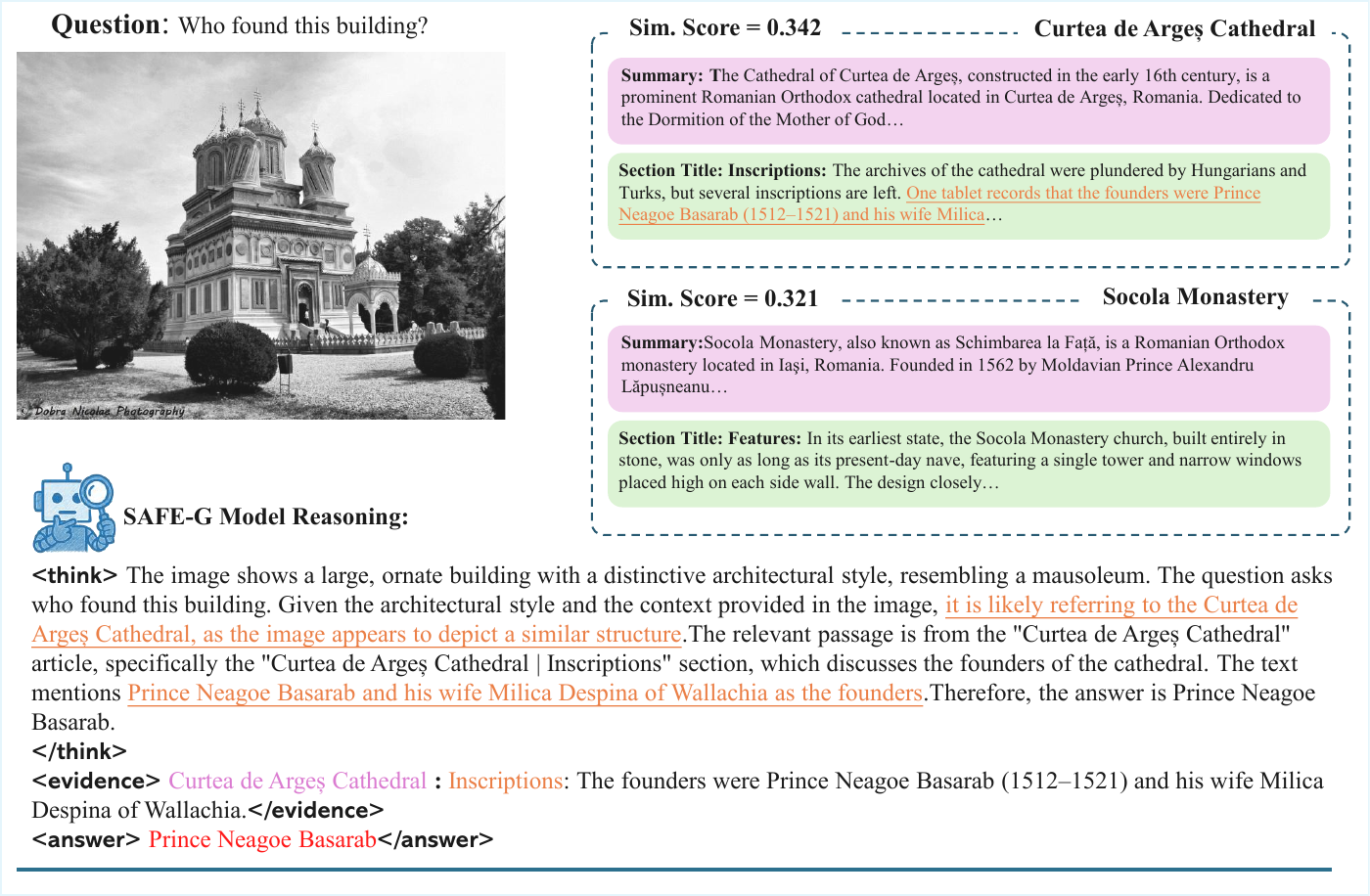}
  \end{subfigure}
  \caption{Qualitative case studies on E-VQA (left) and InfoSeek (right). SAFE-G corrects retrieval bias via graph propagation and precisely localizes target evidence sections, with answers faithfully grounded in the cited evidence.}
  \label{fig:case_study}
\end{figure*}

\begin{table}[t]
  \centering
  \caption{Comparison of training paradigms on E-VQA and InfoSeek, isolating the contribution of RL over zero-shot inference and supervised fine-tuning (SFT).}
  \label{tab:sft_vs_rl}
  \setlength{\tabcolsep}{4pt}
  \renewcommand{\arraystretch}{1.1}
  \resizebox{\linewidth}{!}{
  \begin{tabular}{l l c c c c}
    \toprule
    \multirow{2}{*}{\textbf{Method}} & \multirow{2}{*}{\textbf{Training}} & \multicolumn{1}{c}{\textbf{E-VQA}} & \multicolumn{3}{c}{\textbf{Infoseek}} \\
    \cmidrule(lr){3-3} \cmidrule(lr){4-6}
     & & Single-Hop & Unseen-Q & Unseen-E & All \\
    \midrule

    SAFE-G & Zero-shot & 34.8 & 33.8 & 35.4 & 34.9 \\

    SAFE-G & SFT & 40.4 & 41.7 & 42.9 & 42.3 \\

    \rowcolor{blue!5}
    \textbf{SAFE-G (Ours)} & \textbf{RL} & \textbf{45.1} & \textbf{46.3} & \textbf{47.1} & \textbf{46.7} \\

    \bottomrule
  \end{tabular}}
\end{table}

To clarify the source of performance gains, we compare three training paradigms under identical graph-retrieved evidence: zero-shot inference, supervised fine-tuning (SFT), and our GRPO-based RL, all using the same training set size. As shown in Tab.~\ref{tab:sft_vs_rl}, the zero-shot variant achieves 34.8\% on E-VQA and 34.9\% on InfoSeek, confirming that high-quality graph retrieval alone provides a strong foundation. SFT further improves to 40.4\% and 42.3\% respectively, demonstrating that training with evidence-grounded annotations helps the model better utilize the retrieved context. However, RL outperforms SFT by \textbf{4.7} points on E-VQA and \textbf{4.4} points on InfoSeek. This gap arises because SFT trains the model to mimic fixed annotated reasoning trajectories, which may be suboptimal or inconsistent. In contrast, GRPO explores diverse generation trajectories and reinforces only those that simultaneously cite correct evidence and produce accurate answers, allowing the model to discover more effective reasoning strategies beyond what static supervision can provide.

\subsubsection{Analysis of Training Dynamics}
\label{sec:train_analysis}
We analyze the training dynamics of the reward functions as shown in Fig.~\ref{fig:app_reward}. In the initial phase, the accuracy reward $r_{\text{acc}}(o_i)$ remains low as the model struggles with precise document retrieval and evidence faithfulness. This is primarily due to our strict evidence-gating mechanism: when $r_{\text{sec}}=0$, the accuracy reward is suppressed regardless of answer correctness, forcing the model to first master evidence localization before receiving positive accuracy feedback. As training progresses, all three metrics—$r_{\text{acc}}$, $r_{\text{sec}}$, and the overall reward—increase synchronously, indicating that the model gradually masters the complete reasoning chain: identifying the visual entity, mapping it to the correct document, and pinning down the specific evidence section. By the final stage, the reward curves converge and stabilize, validating that our carefully designed reward mechanism effectively guides the model to perform evidence-grounded answer generation.


\subsubsection{Sensitivity to Training Data Size}

To investigate the sensitivity of SAFE-G to the amount of RL training data, we finally vary the training set size from 2K to 6K samples and report E-VQA performance for both model scales in Tab.~\ref{tab:data_sensitivity}. Both SAFE-G-3B and SAFE-G-7B exhibit clear diminishing returns: SAFE-G-7B improves from 43.5\% at 2K to 45.1\% at 4K, but plateaus beyond that point (44.9\% at 5K, 45.3\% at 6K). A similar trend holds for SAFE-G-3B, which reaches near-peak performance at 5K. This efficiency stems from the nature of RL post-training: rather than memorizing question-answer patterns, RL teaches the model how to ground its reasoning in retrieved evidence and extract the correct answer from external context. Once this reasoning capability is acquired, additional training data yields diminishing returns. These results confirm that SAFE-G is highly sample-efficient, and all main experiments in this paper adopt the 4K setting as the default.



\begin{table}[t]
  \centering
  \caption{Sensitivity of SAFE-G to RL training data size on E-VQA (Single-hop), evaluated at both 3B and 7B model scales with training set sizes ranging from 2K to 6K samples.}
  \label{tab:data_sensitivity}
  \setlength{\tabcolsep}{5pt}
  \renewcommand{\arraystretch}{1.1}
  \begin{tabular}{l l c c c c c}
    \toprule
    \textbf{Method} & \textbf{Model} & \textbf{2K} & \textbf{3K} & \textbf{4K} & \textbf{5K} & \textbf{6K} \\
    \midrule
    \rowcolor{blue!5}
    SAFE-G & Qwen2.5-VL-3B & 37.2 & 38.4 & 39.2 & 39.7 & 39.5 \\
    \rowcolor{blue!5}
    SAFE-G & Qwen2.5-VL-7B & 43.5 & 44.3 & 45.1 & 44.9 & 45.3 \\
    \bottomrule
  \end{tabular}
\end{table}

\subsection{Case Study}


To qualitatively evaluate SAFE-G, we present representative cases from E-VQA and InfoSeek in Fig.~\ref{fig:case_study}. As shown in Fig.~\ref{fig:case_study} (left), the Stage~1 retriever initially ranks a visually similar distractor document higher than the gold document. SAFE-G's structure-aware graph retrieval then propagates multimodal relevance across entity-linked sections, re-ranking the candidate set and successfully elevating the correct evidence to the top. The model subsequently constructs a coherent reasoning chain strictly grounded in the retrieved section, producing the correct answer. This demonstrates SAFE-G's ability to overcome retrieval bias through fine-grained visual-textual consistency verification.

As shown in Fig.~\ref{fig:case_study} (right), the InfoSeek question requires locating a specific piece of information buried deep within a long Wikipedia article (``Inscriptions'' section). SAFE-G's PPR-based propagation precisely identifies this section by aggregating cross-passage structural cues. The evidence-grounded reward then ensures the model explicitly cites the section in \texttt{<evidence>} before deriving the final answer, preventing the model from relying on parametric priors. Together, these cases validate that coupling training-free graph retrieval with evidence-gated RL effectively suppresses hallucinations and anchors responses to verified knowledge.

\section{Conclusion}
In this paper, we presented SAFE-G, a unified framework designed to address the critical challenges of retrieval noise and ungrounded hallucinations in Knowledge-based VQA. By integrating a structure-aware graph retrieval mechanism with a faithfulness-driven reinforcement learning strategy, our approach effectively isolates precise evidence without costly re-ranker training, and enforces strict adherence to the retrieved context during generation. Specifically, our training-free graph retrieval leverages Personalized PageRank with multimodal priors to aggregate cross-passage evidence, while our evidence-grounded GRPO strategy compels the model to anchor its reasoning in verified knowledge by coupling answer correctness with evidence selection. Extensive experiments on Encyclopedic-VQA, InfoSeek, and OK-VQA demonstrate that SAFE-G establishes a new state-of-the-art across benchmarks, outperforming existing baselines by substantial margins.

\bibliographystyle{ACM-Reference-Format}
\bibliography{sections/references}








\end{document}